\pdfoutput=1
\documentclass[11pt,a4paper]{article}
\usepackage{natbib}
\usepackage[margin=25mm]{geometry}

\usepackage{xcolor}		%
\definecolor{darkblue}{rgb}{0, 0, 0.66}
\definecolor{darkred}{rgb}{0.875, 0, 0}
\definecolor{darkgreen}{rgb}{0, 0.5, 0}

\usepackage{placeins}

\newif\ifwithappendix
\withappendixfalse

\newif\ifappendixshown
\appendixshowntrue

\usepackage[T1]{fontenc}
\usepackage[utf8]{inputenc}
\usepackage[textsize=tiny]{todonotes}
\usepackage{graphicx}
\usepackage{booktabs}
\usepackage{latexsym}

\usepackage{footnote}
\makesavenoteenv{tabular}
\makesavenoteenv{table}
\makesavenoteenv{table*}

\newcommand\minput[1]{%
  \input{#1}%
  \ifhmode\ifnum\lastnodetype=11 \unskip\fi\fi}

\usepackage{hyperref}
\usepackage{xstring}

\newcommand{\noqa}[1]{}
\newcommand{\noqall}[1]{}

\usepackage{caption}
\usepackage{amsmath}

\usepackage{tikz}[decoration]
\usepackage{braket} %
\usetikzlibrary{decorations.pathreplacing} %

\hypersetup{colorlinks=true,citecolor=darkgreen, linkcolor=darkred, urlcolor=darkblue}
\title{Oddballness: universal anomaly detection with language models}

\author{\bf Filip Graliński \\
  Adam Mickiewicz University \\
  \texttt{filipg@amu.edu.pl} \\
  \\
  Snowflake \\
  \texttt{filip.gralinski@snowflake.com} \\
  \and \bf Ryszard Staruch \\
  Adam Mickiewicz University \\
  \texttt{ryszard.staruch@amu.edu.pl} \\
  \and \bf Krzysztof Jurkiewicz \\
  Adam Mickiewicz University\footnote{Work partially done for a thesis
  submitted for the MSc degree at Adam Mickiewicz University.} \\
  \\}

\begin{document}

\maketitle

\begin{abstract}
We present a new method to detect anomalies in texts (in general: in
sequences of any data), using language models, in a totally
unsupervised manner. The method considers probabilities (likelihoods)
generated by a language model, but instead of focusing on
low-likelihood tokens, it considers a new metric introduced in this
paper: oddballness. Oddballness measures how ``strange'' a given token
is according to the language model. We demonstrate in grammatical error
detection tasks (a specific case of text anomaly detection) that oddballness is better than just considering
low-likelihood events, if a totally unsupervised setup is assumed.

\end{abstract}

\section{Introduction}

Not all events with low probability are {\em weird} or {\em oddball}
when they happen.
For instance, the probability of a specific deal in the game of bridge is
extremely low ($p_b = \frac{1}{5.36 \times 10^{28}}$ for each deal).
So every time you are dealt cards in bridge, something unfathomable
happens? Of course not, actually {\em an} event of the very low probability $p_b$
{\em must} happen (with the probability 1!).

Another example, imagine two probability distributions:

\begin{enumerate}
  \item $D_1 = \{p_1 = \frac{1}{100}, p_2 = \frac{99}{100}\}$,
  \item $D_2 = \{p_1 = \frac{1}{100}, p_2 = \frac{1}{100}, \ldots p_{100} = \frac{1}{100}\}$,
\end{enumerate}

Intuitively, $p_1$ is much more oddball in $D_1$ than $p_1$ in $D_2$.

So, how to measure {\em oddballness}? We already know that a low probability is not
enough. Let us start with basic assumptions or axioms of oddballness.
Then we will define oddballness and show their practical usage for
anomaly detection when applied to probability distributions generated
by language models.

\section{Axioms of oddballness}

Let us assume a discrete probability distribution
$D = (\Omega, \mathrm{Pr})$, where $\Omega$ could be finite or countably infinite.
From now on, for simplicity, we define $D$ just as a multiset of
probabilities:

\[ D = \{p_1, p_2, p_3, \ldots\} = \{\mathrm{Pr}(\omega_i) : \omega_i \in \Omega\}. \]

We would like to define an oddballness measure\footnote{{\it Measure}
understood informally, not as defined in measure theory.} for an
outcome (elementary event) of
a given probability $p_i$ within a distribution $D$:

\[ \xi_D(p_i), \xi_D : D \rightarrow [0,1] \]

Let us define some common-sense axioms for oddballness:

\begin{description}
  \item[(O0)] $\xi_D(p_i) \in [0,1]$ -- let us assume our measure is
    from 0 to 1,
  \item[(O1)] $\xi_D(0) = 1$ -- if an impossible event happens, that's
    pretty oddball!
  \item[(O2)] for any distribution $\xi_D(\max \{p_i\}) = 0$
    the most likely outcome is not oddball at all,
  \item[(O3)] $p_i = p_j \rightarrow \xi_D(p_i) = \xi_D(p_j)$ -- all
    we know is a distribution, hence two outcomes of the same probability
    must have the same oddballness (within the same distribution),
  \item[(O4)] $p_i < p_j \rightarrow \xi_D(p_i) \geq \xi_D(p_j)$, if
    some outcome is less likely than another outcome, it cannot be less
    oddball,
  \item[(O5)] (continuity) for any distribution  $D = \{p_1, p_2,p_3,\ldots\}$,
    the function $f(x) = \xi_{D_x}(x)$, where
    $D_x = \{x, p_2 \times \frac{1-x}{1-p_1}, \ldots, p_i \times  \frac{1-x}{1-p_1}, \ldots\}$,
    is continuous -- if we change the probabilities a little bit, the
    oddballness should not change much.
\end{description}

Note that (O2) implies the following two facts:

\begin{description}
  \item[(F1)] $p_i > 0.5 \rightarrow \xi_D(p_i) = 0$, what is more likely
    than 50\% is not oddball at all,
  \item[(F2)] for any distribution
     $D=\{p_1=\frac{1}{N},\ldots,p_N=\frac{1}{N}\}$, $\xi_D(p_i) = 0$ --
     like in the bridge example.
\end{description}

\section{Oddballness measure}

Let us a define a measure that fulfils (O0)-(O5). First, let us define
an auxiliary function:

\[ x^{+} = \max(0, x)\]

\noindent (In other words, this is the ReLU activation function.)

Now let us assume a probability distribution
$D = \{p_1, p_2, p_3, \ldots\}$. Let us define the following
oddballness measure:

\[ \xi_D(p_i) = \frac{\sum_j g((p_j - p_i)^{+})}{\sum_jg(p_j)}, \]

\noindent where $g$ is any monotonic and continuous function
for which $g(0)=0$ and $g(1)=1$.

This measure satisfies the axioms (O0)-(O5).

From now on, we assume the identity function $g(x)=x$ (though, for
instance $x^2$ or $x^3$ can be used as well); the oddballness measure
simplifies to:

\[ \xi_D(p_i) = \sum_j (p_j - p_i)^{+}. \]

Let us check this measure for our distributions $D_1$ and $D_2$ given
as examples:

\begin{itemize}
  \item $\xi_{D_1}(p_1) = 0.98$,
  \item $\xi_{D_1}(p_2) = 0$,
  \item $\xi_{D_2}(p_i) = 0$,
\end{itemize}

Consider another example: $D_3 = \{p_1=0.7, p_2=0.25, p_3=0.05\}$,
then: $\xi_{D_3}(p_1) = 0$, $\xi_{D_3}(p_2) = (0.7-0.25)^{+} + (0.25-0.25)^{+} + (0.05-0.25)^{+} = 0.45$,
$\xi_{D_3}(p_3) = 0.85$.

\section{Oddballness as a complement of probability of probability}

Interestingly, oddballness can be interpreted as the complement of
the probability {\em of a probability}. By probability of a
probability $p_i$ with respect to distribution $D$, or $\pi_D(p_i)$,
we mean the probability that an event of probability $p_i$ (not
necessarily $\omega_i$) happens, with two extra assumptions:

\begin{itemize}
\item all probabilities smaller than $p_i$ are also summed up,
\item for each event $\omega_j$ with probability $p_j > p_i$, we
  assume that it contains a ``subevent'' of probability $p_i$, hence
  for each such event we sum $p_i$ in.
\end{itemize}

It can be shown that

\[ \pi_D(p_i) = 1 - \xi_D(p_i). \]

\begin{figure}
\centering
\begin{tikzpicture}
    \draw (0,0) rectangle (6,4);

    \draw (0,0.2) -- (6,0.2);
    \draw (0,1.2) -- (6,1.2);
    \draw[dashed](0,2.2) -- (6,2.2);

    \draw[decorate,decoration={brace,amplitude=10pt,mirror}] (6,2.2) -- (6,4) node[midway,xshift=27pt] {$\xi_D(p_2)$};

    \draw[decorate,decoration={brace,amplitude=10pt,mirror}] (6,0) -- (6,2.2) node[midway,xshift=27pt] {$\pi_D(p_2)$};

    \draw[decorate,decoration={brace,amplitude=10pt}] (0,0) -- (0,0.2)
    node[midway,xshift=-33pt] {$p_1 = 0.05$};

    \draw[decorate,decoration={brace,amplitude=10pt}] (0,0.2) -- (0,1.2)
    node[midway,xshift=-33pt] {$p_2 = 0.25$};

    \draw[decorate,decoration={brace,amplitude=10pt}] (0,1.2) -- (0,4)
    node[midway,xshift=-33pt] {$p_3 = 0.7$};

\end{tikzpicture}
\caption{Illustration of oddballness $\xi_D$ and ``probability of
  probability'' ($\pi_D$) for event $\omega_2$ of probablity
  $p_2=0.25$ for $D_3 = \{p_1=0.7, p_2=0.25, p_3=0.05\}$}
\label{fig:odd}
\end{figure}
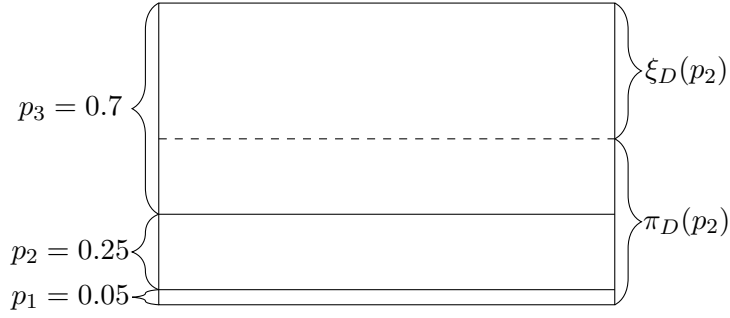

Intuitively, it makes sense: An event is oddball if the probability of
any event happening with similar probability is low. See
Figure~\ref{fig:odd} for an illustration of the relation between
oddballness and probability of probability.

\section{What's the practical use?}

The oddballness measure can be used to detect anomalies or errors, e.g.~in a
text, assuming that we have a good language model. The language model will
give a probability distribution for any word in a text, some words
will be given higher probability (likelihood), some lower. We could mark words
with low probability as suspicious, but sometimes a low-probability
event {\em must} occur. For instance, the distribution for the gap in
the sentence:

\textit{I was born in \dots, a small village}

\noindent should be (for a good language model\footnote{For this
example, an encoder-only model trained on the masked language task
should be assumed, for instance RoBERTa \cite{liu2019roberta}.}) composed of a large number of names, each with a
rather low probability. Hence, like in the bridge example, we should
be not surprised to see a low-probability event. On the other hand, in the
sentence:

\textit{I was born in New \dots City}

\noindent any word other than \textit{York} is pretty unlikely (and
{\em oddball}). Therefore, rather than probability, the oddballness
should be used -- words with oddballness exceeded some threshold
should be marked as suspicious, they are potential mistakes or
anomalies to be checked by humans. This way, we could devise a grammar
checking/proofreading system that is not trained or fine-tuned in a supervised
manner for the specific task of error detection.

The notion of oddballness might not be that useful in the world before
good language models, when usually only static discrete distributions
were assumed. Language models, even for the same text, can generate
vastly different types of probability distributions for each
position:

\begin{itemize}
  \item sometimes the model is almost certain and almost all
    probability will be assigned to one token,
  \item sometimes the model will predict a group of possible tokens
    plus a long tail of less likely tokens,
  \item and sometimes the model is uncertain and the entropy is high.
\end{itemize}

In this paper, we focus on applying oddballness to grammatical error
detection (see Section~\ref{sec:experiments}). Some related (but not
the same) ideas were, however, proposed in the field of log anomaly
detection, as log sequences can be viewed as a modality similar to
natural language. LogBERT by~\cite{guo2021logbert} was trained on, in a
semi-supervised way, on log sequences. During anomaly detection some
tokens are masked and the probability distribution is obtained from
LogBERT for each of them. If the probability of the actual token is
not one of the $K$ highest-likelihood tokens ($K$ is a hyperparameter),
the token is considered anomalous (we will refer to this method as
top$K$ later). LogGPT by~\cite{han2023loggpt} is a
similar idea, but applied to an decoder-only GPT-like architecture,
rather than an encoder-only Transformer, but still the same approach
of considering top$K$ prediction is taken for the anomaly detection
itself, though the model is also fine-tuned specifically for anomaly
detection.

In general, there is a vast body of literature on anomaly or outlier
detection (see, for instance: \cite{scholkopf2001},
\cite{breunig2000lof}, \cite{liu2008isolation}). Oddballness is
different, as it considers only probabilities from a language model
(or any other statistical model) rather than any intrinsic feature of
events in question.

\section{Experiments with error detection}
\label{sec:experiments}

\begin{table}[]
\centering
\caption{Results for the Grammatical Errror Detection FCE Dataset.
  Thresholds tuned with the development set.}\label{tab:fce}
\begin{tabular}{llllll}
\hline
Model
& Method      & Threshold & Dev F0.5       & Test F0.5      &
Submission \\ \hline\hline
Unsupervised methods \\ \hline
GPT2-small                                                                                 & Probability & 0.0002    & 35.00          & 37.74          & \href{https://gonito.net/q/1d49bd435aabd2fab8325b78d0ecfeee3e31ed83}{Link}          \\
GPT2-small                                                                                 & Oddballness & 0.84      & 37.27          & 39.19          & \href{https://gonito.net/q/0341cea8b6e3b3f6223fdad15f39f3f4bef0f647}{Link}         \\ \hline
GPT2-XL                                                                                    & Probabilisty & 0.0001    & 36.00          & 38.86          & \href{https://gonito.net/q/0f4e52985006dfef20a5d21bda799a209c86b860}{Link}         \\
GPT2-XL                                                                                    & Oddballness & 0.85      & 38.17          & 40.52          & \href{https://gonito.net/q/5991373712710569435c502ae4abe197b95edc76}{Link}         \\ \hline
Yi-6b                                                                                      & Probability & 0.0005    & 34.38          & 37.35          & \href{https://gonito.net/q/f037bfb8fec065f3cf90f45a7b41320b1366866d}{Link}         \\
Yi-6b                                                                                      & Oddballness & 0.85      & 36.77          & 39.83          & \href{https://gonito.net/q/5e56d7ba7dbee4230f16f89be1308f5014a76da1}{Link}         \\ \hline
Mistral 7b                                                                                 & Probability & 0.0003    & 33.68          & 36.86          & \href{https://gonito.net/q/fd6e917521b432d4ad7bd4a7bb66841ce0ef7726}{Link}         \\
Mistral 7b                                                                                 & Oddballness & 0.89      & 35.04          & 38.00          & \href{https://gonito.net/q/ed93a2b9ac1edea8da750e2a97e2cc6a67b24b35}{Link}         \\ \hline
RoBERTa Base                                                                               & Probability & 0.005     & 32.63          & 33.62          & \href{https://gonito.net/q/33f13a11288d48022d0f26cbd6f0334fe2015da0}{Link}         \\
RoBERTa Base                                                                               & Oddballness & 0.91      & 33.08          & 34.86          & \href{https://gonito.net/q/284ba7134c35958e6787b9f83e24fb8e0286d6ec}{Link}         \\ \hline
RoBERTa Large                                                                              & Probability & 0.014     & 32.74          & 33.39          & \href{https://gonito.net/q/658ce57a3cbea8e5d04d983beb77aeaf83ace1e1}{Link}         \\
RoBERTa Large                                                                              & Oddballness & 0.84      & 34.33          & 35.78          & \href{https://gonito.net/q/ef813cde8f8702fd77b94fdf45bb24e10ba0939c}{Link}         \\ \hline
\begin{tabular}[c]{@{}l@{}}min(GPT2-XL, \\ RoBERTa Large)\end{tabular}                     & Probability & 0.0001    & 36.88          & 39.31          & \href{https://gonito.net/q/72e9c499a32465e9d95f9b7c6c4acf450c67e632}{Link}         \\
\begin{tabular}[c]{@{}c@{}}max(GPT2-XL, \\ RoBERTa Large)\end{tabular} & Oddballness & 0.89      & \textbf{40.32} & \textbf{43.15} & \href{https://gonito.net/q/9bf7988ff0de7a3c623e41d71dea20b182a61b9d}{Link}         \\ \hline
\hline
Supervised methods & & & & & \\ \hline
\cite{rei-yannakoudakis-2016-compositional}                                                                                 & Bi-LSTM           & -         & 46.00          & 41.10          & -          \\ \hline
\cite{bell}                                                                                & BERT-base          & -         & -              & 57.28          & -          \\ \hline
\cite{kaneko}                                                                               & MHMLA           & -         &                & 61.65          & -
\\ \hline
\cite{yuan-etal-2021-multi}                                                                                 & ELECTRA           & -         & -              & \textbf{72.93}          & -          \\ \hline
\end{tabular}
\end{table}

Table~\ref{tab:fce} presents the results on the FCE
dataset \cite{yannakoudakis-etal-2011-new}. In each case, using the
oddballness value as the threshold gives better results than using the
probability value. All thresholds were adjusted to maximize the F0.5
score on the development set. The maximum oddballness value from the
GPT2-XL and RoBERTa Large \cite{liu2019roberta} models produced the best F0.5 score on the
test set. The result is slightly better than the BiLSTM model by \cite{rei-yannakoudakis-2016-compositional},
which was trained specifically to detect errors in texts, while GPT2-XL and RoBERTa
Large are models which were trained, in a self-supervised manner, on the masked token prediction
task. Although results based on the oddballness value are not competitive with
state-of-the-art solutions, it should be noted that the oddballness
technique {\em does not involve any task-specific fine-tuning}, except
for single-hyperparameter tuning. Also, the texts were written by CEFR B level
students, indicating that they may not be fully proficient in the
language. This could cause the language model to flag not fluent words
as incorrect and thus predict correct words as erroneous. This may also
explain why the smaller GPT2-small model outperforms the much larger
Mistral 7b model. This study demonstrates that the oddballness measure
can yield superior results compared to using probability values for
anomaly detection.

\begin{table}[]
\centering
\caption{Results for the Mistral 7b model on MultiGED-2023 shared task dataset}\label{tab:multiged}
\begin{tabular}{lllll}
\hline                                                                                      Language &Method      & Threshold & Dev F0.5       & Test F0.5      \\ \hline
Czech & TopK & 30      & 41.19          & 38.55      \\
Czech & Probability & 0.002    & 44.34          & 41.90          \\  Czech & Oddballness & 0.84      & 49.16          & 46.61      \\   \hline
German & TopK & 86    & 30.55         & 28.56          \\
German & Probability & 0.001    & 32.53         & 31.36          \\   German & Oddballness & 0.89      & 37.69          & 36.68      \\ \hline
English - FCE & TopK & 200    & 29.14          & 31.60          \\
English - FCE & Probability & 0.00009    & 32.51          & 34.42          \\  English - FCE & Oddballness & 0.90   & 35.07   & 35.86      \\ \hline
English - REALEC & TopK & 380    & 27.62          & 28.10          \\
English - REALEC & Probability & 0.00006    & 31.08          & 31.05          \\ English - REALEC &  Oddballness & 0.95      & 32.89          & 32.09      \\ \hline
Italian & TopK & 18    & 22.76          & 24.55          \\
Italian & Probability & 0.003    & 23.66          & 26.00          \\ Italian & Oddballness & 0.8      & 27.31          & 29.75      \\ \hline
Swedish & TopK & 36    & 35.14          & 33.93          \\
Swedish & Probability & 0.003    & 37.34          & 36.11          \\ Swedish & Oddballness & 0.79      & 40.26          & 38.93      \\ \hline

\end{tabular}
\end{table}

We also tested the Mistral 7b model for multilingual GED datasets used in
MultiGED-2023 Shared Task \cite{volodina-etal-2023-multiged} using the
same approach as in experiments for the FCE dataset. The results in
Table~\ref{tab:multiged} show that for all languages the
oddballness method outperforms the probability method. We also tested
adding the following prompt before each sentence: "An example of a
grammatically correct text in any language that may be out of context:
<example>" to make probability distribution more smooth. The results
in Table~\ref{tab:multigedv2} show that this trick helps in almost
all experiments, but the improvements for the oddballness method are
greater compared to the probability method. Looking at the thresholds
we can also indicate that thresholds for the oddballness value are
more universal compared to the probability thresholds. We also tested the top-$K$ approach. For multilingual GED task it does not provide better results than probability method in any language. The best
solutions for each dataset in the shared task are better compared to
oddballness value results, but again those solutions are trained to
predict incorrect tokens, whereas the oddballness method approach
focus more on predicting spans in texts that are most likely
errorneous without precisely labeling all incorrect tokens.

\begin{table}[]
\centering
\caption{Results for the Mistral 7b model on MultiGED-2023 shared task
  dataset with an additional prompt}\label{tab:multigedv2}
\begin{tabular}{lllll}
\hline                                                                                      Language & Method      & Threshold & Dev F0.5       & Test F0.5      \\ \hline
Czech & TopK & 18    & 41.22          & 39.52          \\
Czech & Probability & 0.002    & 44.26          & 42.70          \\  Czech & Oddballness & 0.85      & 49.68          & 47.75      \\ \hline
German & TopK & 72    & 31.78         & 30.39          \\
German & Probability & 0.0008    & 34.48         & 32.91          \\   German & Oddballness & 0.89      & 39.44          & 39.26      \\ \hline
English - FCE & TopK & 140    & 29.76          & 31.72          \\
English - FCE & Probability & 0.0003    & 32.87          & 34.91          \\  English - FCE & Oddballness & 0.90   & 35.96   & 36.37      \\ \hline
English - REALEC & TopK & 500   & 27.70          & 27.60          \\
English - REALEC & Probability & 0.00008    & 30.44          & 30.18          \\ English - REALEC &  Oddballness & 0.92      & 32.81          & 32.35      \\ \hline
Italian & TopK & 74    & 23.17          & 24.55          \\
Italian & Probability & 0.0008    & 25.26          & 26.56          \\ Italian & Oddballness & 0.92     & 31.66          & 32.62      \\ \hline
Swedish & TopK & 50    & 36.96          & 34.93          \\
Swedish & Probability & 0.002    & 39.34          & 38.26          \\ Swedish & Oddballness & 0.84      & 43.45          & 41.99      \\ \hline

\end{tabular}
\end{table}

\section{Conclusions}

We have showed that using a new metric for anomalous events,
oddballness, is better than just considering low-likelihood tokens, at
least for grammatical error detection tasks. The method based on
oddballness yields worse results than state-of-the-art models heavily
fine-tuned for the task (\cite{Bryant_2023}), but its great advantage is that it can be
used for any language model, without any fine-tuning. This technique
can be applied potentially to anomaly detection in sequences of any
type of data, assuming that a ``language'' model was pre-trained.

\ifwithappendix
\FloatBarrier

\clearpage

\appendix

\input{appendix}
\fi

\bibliography{ms}
\bibliographystyle{abbrvnat}

\end{document}